# A Novel Technique For Grading Of Dates Using Shape And Texture Features

Mohana S.H. and Prabhakar C.J.

Department of Computer Science, Kuvempu University, Shankaraghatta-577451, India

## Abstract

*This paper presents a novel method to grade the date fruits based on the combination of shape and texture features. The method begins with reducing the specular reflection and small noise using a bilateral filter. Threshold based segmentation is performed for background removal and fruit part selection from the given image. Shape features is extracted using the contour of the date fruit and texture features are extracted using Curvelet transform and Local Binary Pattern (LBP) from the selected date fruit region. Finally, combinations of shape and texture features are fused to grade the dates into six grades. k-Nearest Neighbour(k-NN) classifier yields the best grading rate compared to other two classifiers such as Support Vector Machine (SVM) and Linear Discriminant(LDA) classifiers. The experiment result shows that our technique achieves highest accuracy.*

## Keywords

## 1. Introduction

The high dimensional data from the real world are collected using various imaging techniques and collected data are further analyzed in order to develop many applications, which are very much useful for mankind. The main aim of this development is that to duplicate the powers of human vision by electronically perceiving and understanding an image. Computer vision systems are mainly based on imaging techniques such as optical, multispectral and hyperspectral. They have been developed for automatic external quality inspection of agriculture products in many food industries and this is one of the applications, which comes under the computer vision systems. Food and agricultural product grading is done using manually i.e., by the human experts have some limitations such as it is time consuming, tedious, costly, variability and inconsistency. The main aim of the computer vision based approach for food and agricultural products grading is to make the procedure fully automated and to reduce the human intervention. Grading of agricultural products is estimated based on the surface properties such as defects, bruises, irregular shape or size, skin color, skin wrinkles, etc. The grading system employs computer vision techniques to evaluate the surface defects or damages of agricultural products. The damage or defect is usually occurred in agricultural products due to various factors such as in post harvesting, transportation, fungal growth and injury. To improve the customers acceptance rate and also to reduce the cross contamination of agricultural products, we must have to separate these damaged or defected agricultural products from the good one. Appearance is one of the important qualities of fruits and vegetables, which influence their market value based on consumer's preferences. External quality of fruits and vegetables is generally evaluated by considering their color, texture, size, shape, as well as the visual defects.

The most significant physical parameters such as shape, external appearance and size are used for fresh fruits grading. They are characterized by fruit weight, volume, color, weight, diameter and length. Many researchers have introduced varieties of techniques in the field of fruit grading. Some of them are such as apple grading is done by [1]-[8] using their different methodologies, Xu Liming et al. [9] proposed a strawberry grading technique and also many fruits grading techniques have been proposed and adopted for the automated fruit grading systems. Date fruits are usually an oblong brown berry like fruit. Dates have been an important food in desert regions, and are the source of alcohol, syrup, vinegar, and liquor. It is firm, juicy and crunchy with relatively low astringency. The color of the date is one of the significant parameters for post harvest and also for consumers to purchase. There are different varieties of dates are present, in which some varieties have same color and some varieties have different color. These colors in varieties confuse the consumers while purchasing dates and also difficult to differentiate varieties of dates. Another application of dates grading is that growing demand for the supply of best quality dates by the consumers and by date products manufacturing factories. The quality of dried fruits is determined by using hardness, it increases the toughness and chewiness of the fruits. During manufacturing date products such as date syrup and paste, the hard dates can damage the machine components which can reduce the performance of that component for the next coming dates. The admixture process of set of hard dates with the set of soft dates dilutes the buying rate of consumers. Therefore, it is necessary to identify whether the given date is having hardness or semi-hardness or softness it helps to develop a robust date quality evaluation system.

In this paper, we proposed computer vision based date fruits grading algorithm using images of dates to classify or grade the dates into six classes such as hard surface with small size, hard surface with large size, semi hard surface with small size, semi hard surface with large size, soft surface with small size and lastly soft surface with large size, based on fusion of their shape and texture properties. The method begins with reducing the specular reflection and small noise using bilateral filtering. Background removal and fruit part selection is performed using threshold based segmentation. Small holes created by the threshold based segmentation are filled using hole filling approach. This is followed by edge detection using sobel operator to extract a contour of fruit part, and it helps to extract shape features of date. Shape features such as perimeter, area, major-axis, minor-axis, eccentricity and equidiameter are used to represent the shape of the date. LBP based texture information of fruit area is extracted and represented using LBP map. Further, curvelet coefficients of LBP map is used to compute the mean and standard deviation, which represent the texture information in a compact form. Finally, feature level fusion of shape and texture features are used to grade the dates into six grades. The k-NN classifier is employed for date grading because it yields best grading accuracy compared to Support Vector Machine (SVM) and Linear Discriminant Analysis (LDA) classifiers.

The organization of remaining sections of the paper is as follows: The literature review of related work is presented in the section 2. In section 3, theoretical description and mathematical equations of shape features and texture features extraction approaches is given. The proposed approach for date grading is discussed in section 4. Section 5 demonstrates the experimental results. Finally, Section 6 draws the conclusion of this study.

## 2. RELATED WORK

The literature survey reveals that very few papers address grading of dates. D.J Lee et al. [10] proposed a rapid color grading technique for dates quality evaluation using direct color mapping. All datasets were captured on the blue plastic belt it helps for segmentation and also to extract fruit area from the image. The color mapping method maps the colors of RGB values of interest into color indices using a polynomial equation. 3D colors are converted to 1D color indices (3D to 1D color mapping) using the full rank of second-order polynomials. Using color value and their color consistency dates are graded, in which they set the cut-off points of color value and color consistency for each grade. D.J Lee et al. [12] proposed a date grading system using digital reflective near-infrared imaging. They considered two important factors for date grading, which includes fruit size and skin delaminated features for high-quality dates such as Madjools. Based on the size and skin delaminated features the dates are sorted into four grades such as jumbo, extra fancy, fancy and utility. Fruit size is measured using connected component analysis, which calculates the total number of pixels in the fruit region. To improve the fruit area from the segmented image, a Sobel edge detecting operator is applied to extract the delaminated effects on the fruit surface. It detects the edges which are having high contrast or change in intensity in each image. After edge detection, one of the morphological operations such as erosion is applied to remove the few outliers. Small binary mask is logically ANDed with the eroded image to get the final delaminated areas from the date. Finally, the date grading is done by using the size of the date measured in millimeter (mm) and the date skin delaminate is measured in percentage (%). Yousef A.O. et al. [15] developed a computer vision based system for date fruit grading. The system used total 1860 datasets which includes 3 different grades of date fruits such that each grade consists of 620 samples. Simple threshold based segmentation is used to extract fruit area from the background. After segmentation, edges are extracted using Sobel edge operator. The external quality factors such as flabbiness, size, shape, intensity and defects are used as features. Classification is performed based on the extracted features using backpropagation neural network classifier.

A Manickavasagan et al. [13] proposed a technique for grading of dates using RGB color imaging. Three varieties of dates such as Fard, Khalas and Nagal fruits are used in their research. In the first step, simple threshold based segmentation is performed to remove the background and also morphological operations such as hole filling and dilation is performed to improve the results of segmentation. Features are extracted from the segmented fruit image, 13 features are extracted from each color channel and in total 39 features are extracted and used for the classification task. Features are extracted using Gray Level Co-occurrence Matrix (GLCM). Classification is done using Linear Discriminant Analysis (LDA) for three class model (hard, semi-hard and soft) and for two-class model (soft and hard) using Stepwise Discriminant Analysis (SDA) classifier. For three classes model the mean classification accuracy from 68% to 86% and similarly for two-class model mean classification accuracy from 83% to 96% achieved. D. Zhang et al. [14] proposed an approach for date quality evaluation based on color distribution analysis and back projection approach. Fruits are placed on the blue color plastic belt, and background is subtracted using threshold based segmentation. Logical OR operation is performed on Red and blue masks to create a combined mask to reduce the holes and missing areas of the fruit. Blue channel does not have much discrimination power for grading, so they used blue background in their experiment. Experiments conducted on 353 Medjool date samples, it consists of four classes such as class 1(dark red), class 2 (light red), class 3 (orange) and class 4 (yellow). The histograms of red and green channels are created, and normalized. Each pixel in the segmented area (fruit area) is back projected using back projection matrix and index value is assigned to it based on the weighted-average approach. A unique refinement is done for the back project matrix to get the index value. Finally, the color grading is performed based on the average of color indexes for each date.

D. Zhang et al. [11] proposed an approach for evaluation of date quality using short-wave infrared imaging. Image acquisition is done by using short-wave infrared (SWIR) cameras, from the captured datasets it is observed that the dark pixel represents the tight skin and light pixel represents the delaminated skin. The background is subtracted using simple threshold based segmentation approach. Ellipse fitting approach is adopted on the segmented fruit and fruit size is measured based on the major axis length. The fruit grading is performed based on the measurement of fruit size and the percentage of delaminated skin. Based on the image binarization and ellipse fitting, they measured the size of date fruit. In fruit testing phase, the delaminated skin is separated from the normal fruit based on the threshold calculated from the histogram similarity. The training stage includes the gray-scale histogram normalization and fruit skin threshold for each prior quality class. Normalized histogram of the test fruit is compared with the four predefined classes. The fruit size is measured between the major axis of the ellipse fitted on the fruit and two intersections of the fruit boundary. To separate delaminated skin from normal fruit, skin threshold is calculated according to histogram similarity is applied to the test fruit. For grade determination, four levels of delamination are compared based on the delamination percentage. The delamination percentage is less than 10% belongs to class 0, between 10% and 25% for class 1, for class 2 percentage range from 25% to 40% and finally the delamination percentage greater than 40% belongs to the class 3. The skin delamination percentage is estimated based on the Sum of Absolute Difference (SAD) approach, in which the difference between the normalized histogram i.e., the measure of gray level distribution distance difference of test fruit is compared with the predefined class. In training stage, 40 date samples (10 from each class) are selected and used for training. Total 1200 date samples (300 from each class) are selected and used for testing. The accuracy of their approach was 95% for jumbo and extra fancy classes and similarly, for fancy and confection classes 98% accuracy has been achieved.

## 3. OUR APPROACH

The proposed dates grading approach comprises the steps such as pre-processing, fruit part selection, shape and texture features extraction and finally grade the dates using supervised classifier. The detailed description of these steps is explained in the following sections. The schematic diagram of proposed approach is shown in the Figure 1.

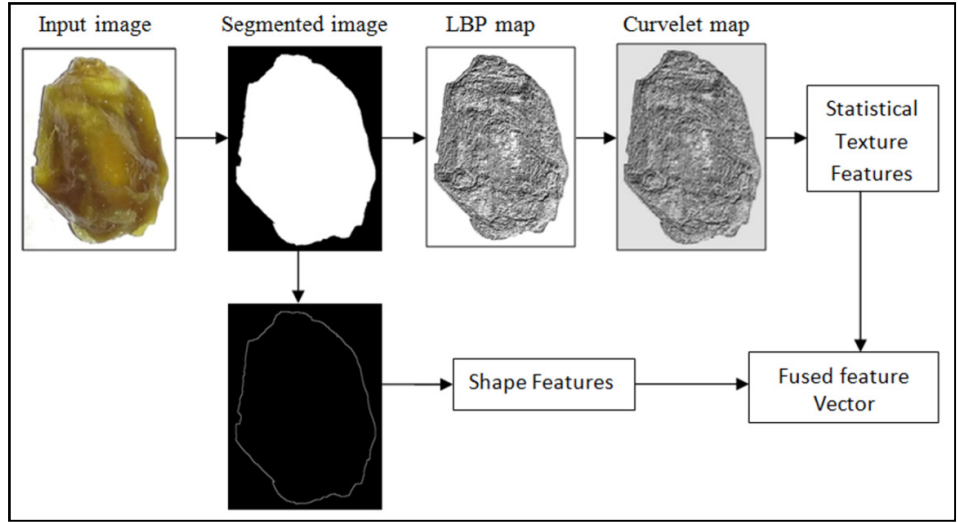

Figure 1. Schematic diagram of proposed methodology

### 3.1 Pre-processing

The pre-processing stage includes the steps such as a bilateral filter is used to remove the specular reflection from the given input image. In the second step threshold based segmentation is performed to extract the date fruit region. This is followed by the hole filling operation, which is used to remove the small holes that are created in the threshold based segmentation step. Finally, the contour of the fruit area is extracted using sobel edge detection operator.

Qingxiong Yang et al. [21] proposed an approach for real time specular highlight removal using a bilateral filter. A bilateral filter is a non-linear smoothing filter in which it preserves edges, reduces noise and specular reflection information from the given images. In the image, the intensity at each pixel is replaced with a weighted average of the intensity values from neighboring pixels. The weights are estimated based on Gaussian distribution, and the weights are not only dependent on the Euclidean distance of neighbor pixels, but also on the radiometric differences such as color intensity difference, depth distance and so on. The main advantage of this filter is that, it preserves sharp edges based on the average weights of the neighbor pixels which are estimated.

Chromaticity is defined as the fraction of color component c

$$\sigma_c = \frac{J_c}{\sum_{c \in \{r,g,b\}} J_c} \ , \tag{1}$$

where $J$ is the reflected light color.

$$\sigma_{min} = min(\sigma_r, \sigma_g, \sigma_b), \tag{2}$$

$$\sigma_{max} = max(\sigma_r, \sigma_g, \sigma_b), \tag{3}$$

Approximate diffuse chromaticity $\lambda_c$ is calculated as

$$\lambda_c = \frac{\sigma_c - \sigma_{min}}{1 - 3\sigma_{min}}, \tag{4}$$

$$\lambda_{max} = max(\lambda_r, \lambda_g, \lambda_b), \tag{5}$$

Using the approximated maximum diffused chromaticity defined in the above equation to supervise the smoothing operation, the filtered maximum chromaticity $\sigma_{max}$ can be computed as

$$\sigma_{max}^F(p) = \frac{\sum_{q \in \Omega} F(p,q) G(\lambda_{max}(p), \lambda_{max}(q)) \sigma_{max}(q)}{\sum_{q \in \Omega} F(p,q) G(\lambda_{max}(p), \lambda_{max}(q))}, \tag{6}$$

where $F$ and $G$ are the spatial and range weighted functions which are typically Gaussian. And $p$ and $q$ are the any two pixels in the image. To exclude the contribution of the specular pixels we compared $\sigma_{max}$ and $\sigma_{max}^F$ and take the maximum value as

$$\sigma_{max}(p) = max(\sigma_{max}, \sigma_{max}^F(p)). \tag{7}$$

### 3.2 Shape Features

The shape features are extracted using the contour of the date fruits. In our approach, we used six shape metrics which includes area, perimeter, major axis length, minor axis length, eccentricity and equidiameter. The mathematical description of these metrics is given below:

### Area (A)

The area of the fruit region is calculated using the total number of pixels which are present inside the fruit region, which describes the area of that region. It is estimated as

$$A = N_T, \tag{8}$$

where $N_T$ is the total number of pixels within the contour.

### Perimeter (P)

The pixels which are present on a boundary of the region define the perimeter of the region based on the contour. The perimeter is calculated as, the distance between each adjoining pair of pixels on the border of the region.

$$P = N_p, \tag{9}$$

where $N_p$ is the total number of pixels around the boundary.

### Major-axis length (MAJL)

It is calculated by using the maximum diameter of the shape, which holds the number of pixels in that longest diameter of the ellipse.

$$MAJL = max(D), \tag{10}$$

where $D$ is the diameter.

**Minor-axis length (MINL)**

It is calculated by using the minimum diameter of the shape, which holds the total number of pixels in that shortest diameter of the ellipse.

$$MINL = min(D),\qquad(11)$$

where $D$ is the diameter.

**Eccentricity (E)**

The eccentricity is the ratio of the distance between the focal points of the ellipse and its major axis length. The value is ranges from 0 to 1 and similarly, an ellipse whose eccentricity is 0 is actually a circle, while an ellipse whose eccentricity is 1 is a line segment.

$$E = \sqrt{(a^2 - b^2)}/a = \{0 \le e \le 1,\qquad(12)$$

where $a$ is the major axis and $b$ is the minor axis.

**Equidiameter (ED)**

It is a type of diameter in which it specifies the diameter of a circle with the same area as the region. It is computed as

$$ED = \sqrt{(4 * A/\pi)},\qquad(13)$$

where A is area represented using pixels.

Table 1. Extracted shape feature values of some sample dates

| Grade | A | P | MAJL | MINL | E | ED |
|---|---|---|---|---|---|---|
| Soft_Small | 14278 | 2335 | 804 | 402 | 0.86 | 134.86 |
| Soft_Large | 16693 | 2708 | 892 | 505 | 0.82 | 145.78 |
| Semi_Hard_Small | 14699 | 2495 | 676 | 510 | 0.65 | 136.80 |
| Semi_Hard_Large | 16089 | 2557 | 726 | 580 | 0.60 | 143.12 |
| Hard_Small | 17004 | 2906 | 904 | 557 | 0.78 | 147.14 |
| Hard_Large | 18302 | 3111 | 967 | 569 | 0.80 | 152.65 |

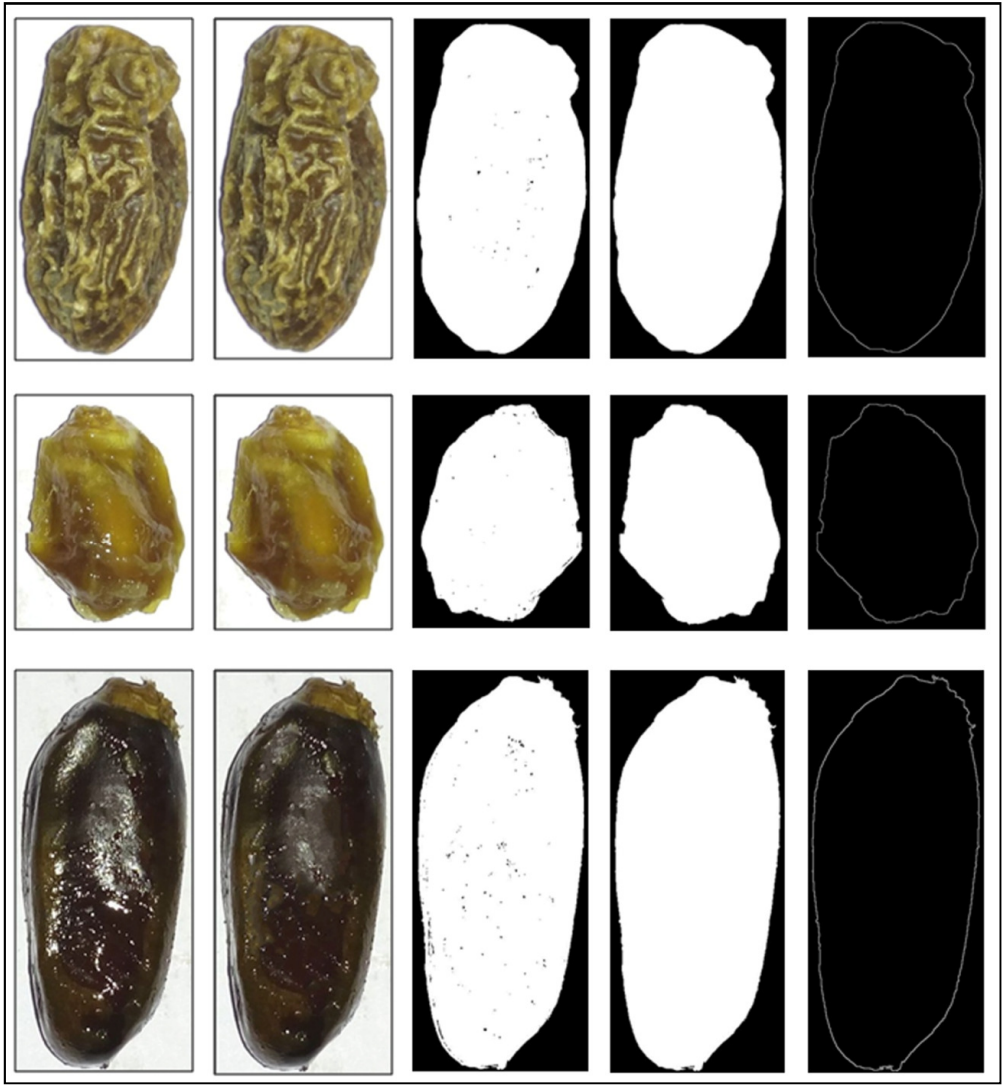

Figure 2. First column: original image, Second column: specular reflection removed image, Third column: segmented results of second column image, Fourth column: results of hole filling operations of third column image, Fifth column: contour of the fruit area

## 3.3 Texture Features

The dates are having smooth variation surface and exhibit unique texture. Since, the dates surface exhibit texture property, which can be further used for differentiating the various grades such as soft, hard and semi-hard dates. The hard skin dates have much transition compared to soft skin dates. Similarly, the semi-hard dates are having less transition compared to hard skin dates. We used LBP operator to compute the LBP map, which contains the texture information of fruit surface. Further, the curvelet coefficients of the texture map are used to extract statistical features such as mean and standard deviation.

Local binary pattern (LBP) is a texture operator, and it is used to extract the texture information from the given input image. Ojala T. et al. [18] proposed the LBP texture operator, and they conducted experiments to measure texture information using this LBP operator and texture classification is performed based on the distribution of texture features. Ojala et al. [19] extended the LBP operator for multi-resolution gray scale and rotation invariant texture classification. A pixel in the given image, an LBP code is computed by comparing the pixel with its neighbours:

$$LBP_{P,R} = \sum_{p=0}^{P-1} s(g_p - g_c) \, 2^p, s(x) = \begin{cases} 1, x \geq 0 \\ 0, x < 0 \end{cases} ,\qquad(14)$$

where $g_c$ is the gray value or intensity of the center pixels, $g_p$ is the gray value or intensity of its neighbors, p is the total number of neighbors are involved in this task, and R is the radius or distance of the neighborhood from the center pixel.

The $U$ value of an LBP pattern is defined as the number of spatial transition i.e., bitwise 0/1 changes in that pattern

$$U(LBP_{P,R}) = |s(g_{P-1} - g_c) - s(g_0 - g_c)| + \sum_{p=1}^{P-1} |s(g_p - g_c) - s(g_{p-1} - g_c)| ,\qquad(15)$$

The patterns which are having limited transition or discontinuities ($U \leq 2$), those patterns are belongs to uniform LBP patterns. In practice, the uniform patterns are denoted as "u2". The mapping from $LBP_{P,R}$ to $LBP_{P,R}^{u2}$, which has $P * (P - 1) + 3$ distinct output values using lookup table.

To achieve rotation invariance, a locally rotation invariant pattern is defined as

$$LBP_{P,R}^{riu2} = \begin{cases} \sum_{p=0}^{P-1} s(g_p - g_c), & if \ U(LBP_{P,R}) \leq 2 \\ P + 1, & otherwise \end{cases} ,\qquad(16)$$

The Figure 3 shows the result of LBP operator on various types of dates. The hard, semi hard and soft dates are shown row-wise respectively. The LBP histogram plotted for these grades demonstrates that the texture information can be used for discriminating the dates based on compact representation of texture information.

## 3.4 Curvelet Transform

Curvelet transform is a non-adaptive approach for object representation with multi-scales. It is an extension of wavelet transform approach, and it is becoming popular in many fields, namely in the computer vision and pattern recognition. Candes et. al., [20] proposed curvelet transform, in which it has a highly redundant dictionary which can provide sparse representation of signals that have edges along the regular curve. Candes et. al., [22] in 2006 they redesigned the curvelet transform, and it was re-introduced as Fast Discrete Curvelet Transform (FDCT). The second version of curvelet transform overcomes the limitations of the previous version i.e. it has fewer redundant and also works faster compared to its first version. It is defined in both discrete and continuous domains and also for higher dimensions. 2D FDCT is used for image based feature extraction. The curvelet coefficients values are determined by using the idea is that, how they are aligned in the real image. The curvelet is aligned more accurately with a given curve in an image, and the coefficient values are higher. In curvelet transform, the edge discontinuity is approximated in a better way compared to wavelets. Curvelets overcome the limitations of wavelets such as curved singularity representation, limited orientation and absence of anisotropic element. FDCT is implemented in two different approaches, i.e., one is via USFFT (Unequally Spaced Fast Fourier Transform) and another one is via Wrapping. Both approaches are linear and Cartesian array is given as input and which provide discrete coefficients as output. Both are differed in implementation phase in which the choice of spatial grid to translate curvelets at each scale and angle. FDCT via wrapping is used in our approach because it is a fastest curvelet transform, which is currently available to estimate the curvelet coefficients for feature extraction. The algorithm of FDCT wrapping is described below:

The 2D discrete Fourier transform of $f[t_1, t_2]$ is denoted as $\hat{f}[n_1, n_2]$. Let $U_j(\omega)$ is a localizing window and $\tilde{U}[t_1, t_2]$ is supported on a rectangle of length $L_j$ and width $W_j$ [22].

$$P_j = \{(n_1, n_2) : n_{1,0} \leq n_1 < n_{1,0} + L_j, n_{2,0} \leq n_2 < n_{2,0} + W_j\}, \tag{17}$$

Implementation steps of FDCT via wrapping [13]

1. Apply the 2D Fast Fourier Transform (FFT) and obtain Fourier samples $\hat{f}[n_1, n_2]$, $-n/2 \leq n_1, n_2 < n/2$
2. The product $\tilde{U}_{j,l}[n_1, n_2] \cdot \hat{f}[n_1, n_2]$ is estimated for each scale $j$ and angle $l$
3. Wrapping is performed on this product around the origin and obtain

$$\hat{f}_{j,l}[n_1, n_2] = W(\tilde{U}_{j,l}\hat{f})[n_1, n_2], \tag{18}$$

   where $n_1$ and $n_2$ is ranges from $0 \leq n_1 < L_j$ and $0 \leq n_2 < W_j$
4. Apply the inverse 2D FFT to each $\hat{f}_{j,l}$ and finally we get the corresponding discrete coefficients for each $\hat{f}_{j,l}$.

In our experiment, we used only curvelet coefficients of approximate sub band image to extract texture features. The statistical measures such as standard deviation and mean are estimated from the coefficients of an approximated sub band of the resultant LBP map of a date fruit. These features represent the texture information and it is denoted as $V_T$ (Texture feature vector). Let $I(p, q)$ be the approximate sub band image, the resulting feature vector is

$$V_T = \{\mu, \sigma\},$$

where $\mu$ and $\sigma$ are the mean and standard deviation of the approximated sub band respectively. The statistical texture features are estimated as follows

$$\mu = \frac{1}{M \times N} \sum_{p=1}^{M} \sum_{q=1}^{N} I(p, q), \tag{19}$$

$$\sigma = \left\{ \frac{1}{M \times N} \sum_{p=1}^{M} \sum_{q=1}^{N} [I(p, q) - \mu]^2 \right\}^{1/2}, \tag{20}$$

where $M$ and $N$ are the size of the approximate sub band image $I(p, q)$.

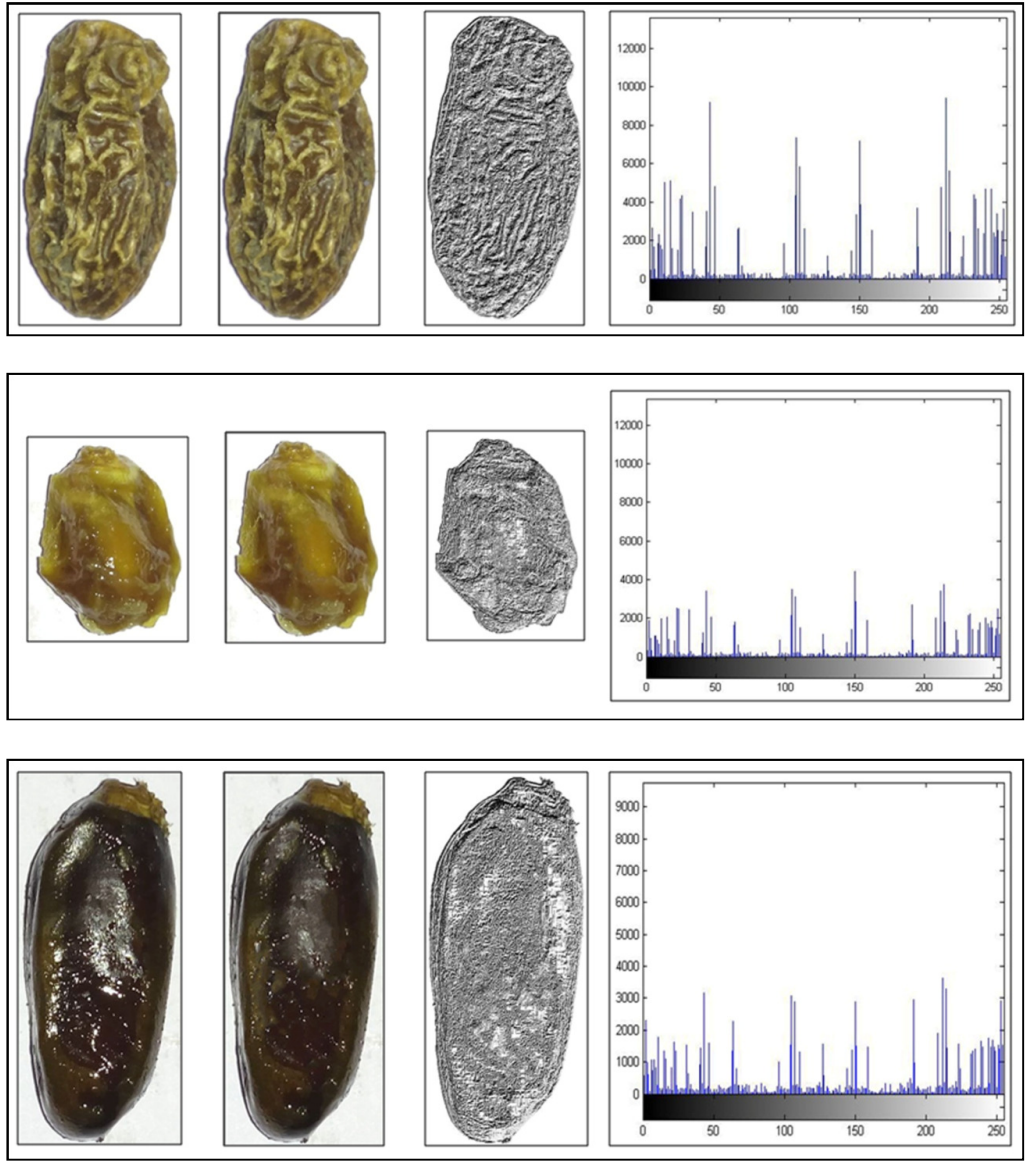

Figure 3. Results of LBP operator: First column: original RGB image, Second column: Specular reflection removed image, Third column: LBP map of second column image, Fourth column: corresponding LBP histogram.

## 3.5 Similarity Measure

The two feature vectors namely texture feature vector ($V_T$) and shape feature vector ($V_S$) is concatenated and is denoted as $V_{ST} = cat(V_T, V_S)$. The similarity between the test image feature vector and reference image vector is computed using

$$E_d(t,r) = \sqrt{\sum_{i=1}^{n}(t_i - r_i)^2},$$ (21)

where $t$ is the test vector , $r$ is the reference vector and n is the size of the vector.

Table 2. Texture features of some samples of dates

| Grade | Mean | Standard Deviation |
|---|---|---|
| Soft_Small | 1021 | 319 |
| Soft_Large | 1083 | 325 |
| Semi_Hard_Small | 1139 | 332 |
| Semi_Hard_Large | 1167 | 345 |
| Hard_Small | 945 | 362 |
| Hard_Large | 987 | 378 |

## 4. EXPERIMENTAL RESULTS

We have conducted the experiments to evaluate the proposed method for dates grading. Since, there is no benchmark database of dates is available, we have created our own date database for experimentation purpose. Image acquisition device is composed of a resolution 5.0 mega pixel (2560 x1920). The images of the selected 960 dates were captured in indoor environment with slight variation in lighting condition. While acquiring the images of data samples, we used the white background, which is convenient for fruit region extraction. The data samples consist of six grades of dates, which were graded by human experts. Among the above-said data samples, we selected randomly 50% (80 samples from each grade) of the date samples for training purpose, and the remaining 50% (80 samples from each grade) samples are used for testing purposes. In order to eliminate the intersection (duplication) of training and testing sets, drop-one-out approach is used in this work i.e., a date used for training is excluded from the testing set.

The experiment procedure is as follows: we separate the fruit area from the background and extract the shape and texture features of the fruit region. As we described in the previous sections, we extract and store the shape and texture features in the database. The experiments are conducted for individual feature vectors and also for fused feature vectors to show the significance of feature fusion approach compared to individual feature vectors. The feature vectors such as texture ($V_T$) and shape ($V_S$) vectors are concatenated to form a joint feature vector ($V_{ST}$), which represent the combination of texture and shape features of date fruit. Here linear basis strategy is used for fusion. The fused feature vectors of six date grades are stored independently in the database. To reduce the time complexity of the grading task, average of feature vectors of the grade samples are considered and treated as reference models in order to grade the dates.

The performance of three supervised classifiers is analyzed in order to find suitable classifier for dates grading. The training and testing sample size is same as mentioned above and conducted experiments using classifiers such as Support Vector Machine (SVM), k-Nearest Neighbor (k-NN) and Linear Discriminant Analysis (LDA). The radial basis kernel function is used for SVM classifier and k values of k-NN classifiers are varied in order to find suitable value. Table 3 shows the comparison of dates grading results for three classifiers and it can be observed that, k-NN classifier (for k=4) yields highest grading rate and this classifier suits for our proposed date grading approach. Also, we demonstrate the effect of fusion of features. From the Table it is observed that features fusion increases the grading rate of dates compared to individual features. This demonstrates the combinations of features are having much discrimination power, and also they have an ability to increase the grading accuracy.

Table 3. Comparison of performance of different classifiers based on grading accuracy (%)

| Classifier | Texture features | Shape features | Texture + Shape features |
|---|---|---|---|
| k-NN (k=4) | **91.37** | **88.95** | **96.45** |
| SVM (RBF kernel) | 90.33 | 87.29 | 91.66 |
| LDA (Multiclass) | 88.75 | 86.45 | 89.58 |

Table 4 shows the results of date grading accuracy for six grades using k-NN classifier due to its ability to yield highest classification accuracy. From the results, it is observed classification accuracy for semi_hard_small grade is low compared to other grades. The highest classification accuracy is achieved for hard_small grade among all grades considered.

Table 4. Dates grading Accuracy (%) of proposed method using k-NN classifier

| Grade | Texture features | Shape features | Texture + Shape features |
|---|---|---|---|
| Soft_Small | 91.25 | 87.50 | 97.50 |
| Soft_Large | 87.50 | 88.75 | 95.00 |
| Semi_Hard_Small | 91.25 | 86.25 | 93.75 |
| Semi_Hard_Large | 96.25 | 85.00 | 96.25 |
| Hard_Small | 88.75 | 93.75 | 98.75 |
| Hard_Large | 93.25 | 92.50 | 97.50 |
| **Average grading accuracy** | **91.37** | **88.95** | **96.45** |

The table 5 describes the confusion matrix of date grading which includes 80 test samples from each grade. This table demonstrates the results of correct and wrong grading. Some soft dates are wrongly graded as semi hard type and some semi_hard dates are graded as both hard and soft grades. Similarly, a very small percentage of hard dates are falsely graded. This demonstrates that, grading of semi_hard and soft dates are effects on the result of false grading compared to hard dates.

Table 5. Confusion matrix for date grading

| Grade | Soft_Small | Soft_Large | Semi_Hard_Small | Semi_Hard_Large | Hard_Small | Hard_Large |
|---|---|---|---|---|---|---|
| Soft_Small | 78 | 01 | 01 | 00 | 00 | 00 |
| Soft_Large | 01 | 76 | 00 | 03 | 00 | 00 |
| Semi_Hard_Small | 01 | 00 | 75 | 02 | 02 | 00 |
| Semi_Hard_Large | 00 | 01 | 01 | 77 | 00 | 01 |
| Hard_Small | 00 | 00 | 01 | 00 | 79 | 00 |
| Hard_Large | 00 | 00 | 00 | 01 | 01 | 78 |

Table 6 shows the evaluation of experimental results based on True Positive Rate (TPR) and False Positive Rate (FPR) which is obtained using k-NN classifier. The TPR represents the correct grading i.e., the dates are correctly graded based on their prior knowledge such as shape and texture features. Similarly, the FPR represents the wrong grading i.e., the dates are not correctly graded based on the prior knowledge. From table 6, we found that the highest TPR is obtained for hard_small grade and the lowest TPR is obtained for Semi_Hard_Small grade. Finally, the average TPR is 96.45% with small FPR of 3.55%.

Table 6. TPR and FPR for dates grading using k-NN classifier

| Grade | TPR | FPR |
|---|---|---|
| Soft_Small | 78 (97.50%) | 02 (2.50%) |
| Soft_Large | 76 (95.00%) | 04 (5.00%) |
| Semi_Hard_Small | 75 (93.75%) | 05 (6.25%) |
| Semi_Hard_Large | 77 (96.25%) | 03 (3.75%) |
| Hard_Small | 79 (98.75%) | 01 (1.25%) |
| Hard_Large | 78 (97.50%) | 02 (2.50%) |
| **Average rate** | **77.17 (96.45%)** | **2.83 (3.55%)** |

## 5. CONCLUSION

This paper presents a novel approach for dates grading, which is purely based on the shape and texture features. In preprocessing step, we used bilateral filter and threshold based segmentation to reduce specular reflection and fruit region extraction respectively. In order to discriminate the various sized dates, we used six shape metrics. Similarly, based on texture features, we were able to discriminate successfully the dates into soft, hard and semi-hard grades. The shape and texture features fusion is performed and improved performance shows the significance of the fusion approach. Our approach yields best grading accuracy of 96.45% with small error rate of 3.55%. The proposed method can be incorporated for automated dates grading system. It can reduce the human intervention and also time required for grading process.